\documentclass[runningheads]{llncs}
\usepackage{graphicx}
\usepackage{amsmath,amssymb} 

\usepackage{color}
\usepackage{capt-of}

\usepackage{multirow}
\usepackage{subcaption}

\usepackage{wrapfig,lipsum,booktabs}

\usepackage{capt-of}

\begin{document}

\title{Toward Learning a Unified Many-to-Many Mapping for Diverse Image Translation} 

\titlerunning{Image Translation}

\author{Wenju Xu \and
Keshmiri Shawn \and
Guanghui Wang}
%
\authorrunning{Wenju \it{et al.}}
%

\institute{School of Engineering, University of Kansas, Lawrence, KS, USA 66045}

\maketitle

\begin{abstract}
		Image-to-image translation, which translates input images to a different domain with a learned one-to-one mapping, has achieved impressive success in recent years. The success of translation mainly relies on the network architecture to reserve the structural information while modify the appearance slightly at the pixel level through adversarial training. Although these networks are able to learn the mapping, the translated images are predictable without exclusion. It is more desirable to diversify them using image-to-image translation by introducing uncertainties, i.e., the generated images hold potential for variations in colors and textures in addition to the general similarity to the input images, and this happens in both the target and source domains. To this end, we propose a novel generative adversarial network (GAN) based model, InjectionGAN, to learn a many-to-many mapping. In this model, the input image is combined with latent variables, which comprise of domain-specific attribute and unspecific random variations. The domain-specific attribute indicates the target domain of the translation, while the unspecific random variations introduce uncertainty into the model. A unified framework is proposed to regroup these two parts and obtain diverse generations in each domain. Extensive experiments demonstrate that the diverse generations have high quality for the challenging image-to-image translation tasks where no pairing information of the training dataset exits. Both quantitative and qualitative results prove the superior performance of InjectionGAN over the state-of-the-art approaches. 

\end{abstract}




\begin{figure*}[t!]
	\centering
	\begin{subfigure}[t]{0.43\textwidth}
		\centering
		\includegraphics[height=4.2cm]{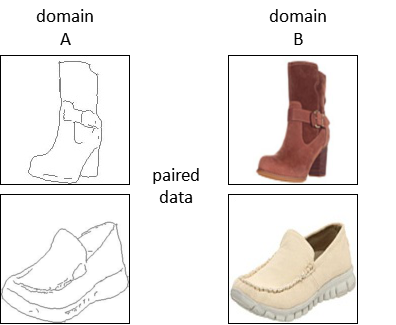}
		\caption{\label{paired}paired dataset}
	\end{subfigure}
	\hspace{5mm}
	\begin{subfigure}[t]{0.45\textwidth}
		\centering
		\includegraphics[height=4.2cm]{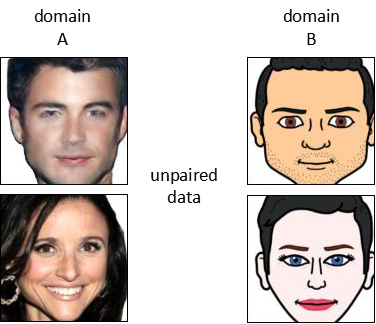}
		\caption{\label{unpaired}unpaired dataset}
	\end{subfigure}
	
	\begin{subfigure}[t]{0.45\textwidth}
		\centering
		\includegraphics[height=2.5cm]{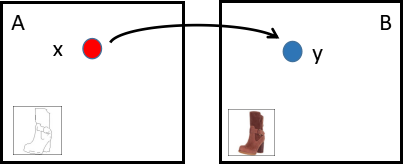}
		\caption{\label{one}one-to-one mapping}
	\end{subfigure}
	\hspace{5mm}
	\begin{subfigure}[t]{0.45\textwidth}
		\centering
		\includegraphics[height=2.5cm]{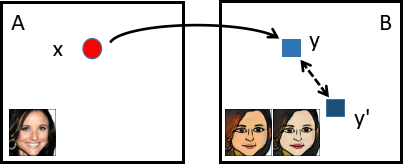}
		\caption{\label{many}one-to-many mapping}
	\end{subfigure}
	
	\begin{subfigure}[t]{0.45\textwidth}
		\centering
		\includegraphics[height=2.5cm]{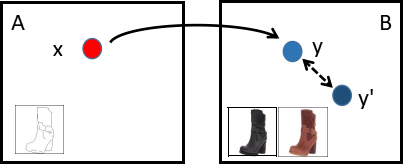}
		\caption{\label{one to many} one-to-many mapping}
	\end{subfigure}	
	\hspace{5mm}		
	\begin{subfigure}[t]{0.45\textwidth}
		\includegraphics[height=2.5cm]{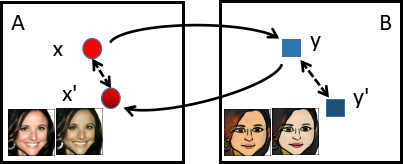}
		\caption{\label{many to many} many-to-many mapping}
	\end{subfigure}
	\caption{\label{Illustration} Illustration of different image-to-image translation tasks.}	
\end{figure*}

\section{Introduction}

Deep neural network models have shown great success in many pattern recognition and computer vision applications \cite{he2018learning,ma2018mdcn}. In recent years, image-to-image translation, which learns an one-to-one mapping by mapping the input images from the source domain to the target domain has attracted a lot of attention. Image translation requires the generated images not only to be with high visual quality but also should be perceptually consistent with the corresponding inputs. Some models \cite{pix2pix2016,zhu2017toward} even learn the one-to-one mapping relying on pairs of corresponding training images in two domains. Nevertheless, recent advances in deep generative models \cite{kim2017learning,liu2017unsupervised} have achieved great success by adversarially learning the one-to-one mapping in real-world applications, such as changing attributes of faces \cite{PENG2018262,choi2017stargan} and transferring the style of images \cite{gatys2016image,li2017diversified}. 

Conditional generative adversary networks (cGANs) are first employed for pixel-level image-to-image translation. Instead of generating images from random noise, the cGANs generate images conditioned on the input images. Its generator learns a one-to-one mapping that translates the input image to the target domain, while the discriminator is trained to discriminate the generated images from those real images in the target domain. It has been shown that these frameworks, such as Pix2pix \cite{pix2pix2016} and CycleGAN \cite{CycleGAN2017}, learn an impressive one-to-one deterministic mapping function which generates high quality samples in the target domain. The StarGAN \cite{choi2017stargan} was proposed to translate face images according to multiple face attributes. It combined one additional classifier with the discriminator so as to learn the mappings in a unified model. However, the generated samples are expectable without exception. A recent popular alternative approach to cGANs is the variational autoencoder (VAE) \cite{kingma2013auto}. It learns a distribution of the encoded latent codes, and matches it to a prior explicit distribution by minimizing a variational lower bound. When combined with GAN, it adversarially learns the latent codes from the observations and generate samples in different domains by either manipulating the encoded latent codes or adding additional labels. This model is usually referred to as the cVAEGAN. In the scenario of image-to-image translation, the BicycleGAN  \cite{zhu2017toward} was the first one that successfully generates diverse samples with high quality. However, it requires paired dataset for learning a random variable which contains the variation in the learned one-to-many mapping. MUNIT model \cite{huang2018munit} learns the mapping in an unsupervised manner. This model generates diverse samples by separately learning the content and the style codes. The diversity comes by enforcing the style codes to be in a normal distribution. Even though these cVAEGAN models achieve diversity in image translation, they are inefficient for multi-domain translation as it needs to train one model for each pair of domains. The differences between these models are demonstrated in Figure \ref{Illustration}.
\begin{figure*}[t!]
	
	\begin{minipage}[b]{1.0\linewidth}
		\centering
		\centerline{\includegraphics[width=13cm]{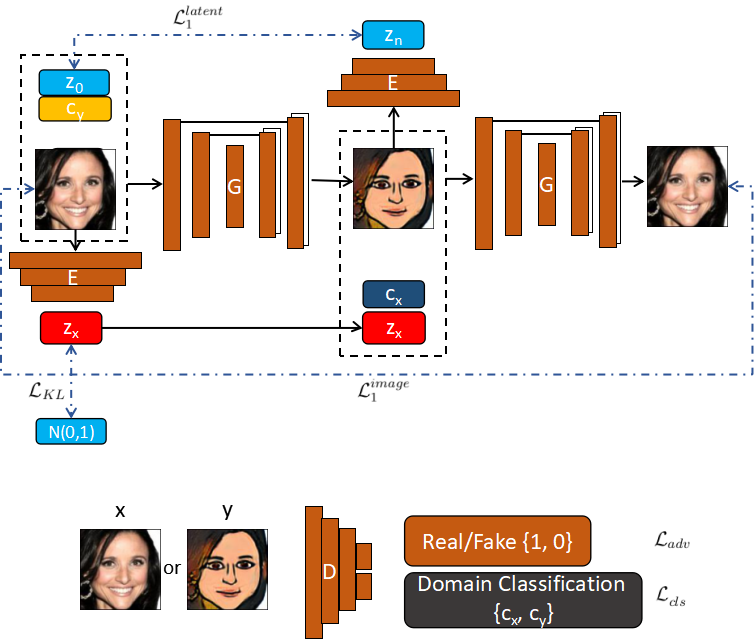}}
	\end{minipage}
	\caption{The proposed network architecture. Our model consists of three modules, a discriminator $D$, a generator $G$, and an encoder $E$. The discriminator $D$ is built upon the patch discriminator. It also classifies the input images to its corresponding domain; the generator $G$ takes in as input containing the image, target domain
		label $c$ and additional latent code $z$; and the encoder $E$ tries to
		encode the image into latent code $z$ as additional information. The lost functions are defined in the text.}
	\label{overview}
\end{figure*}

To overcome the above issues, we propose the InjectionGAN, a deep conditional generative model that unifies the generative adversarial network \cite{goodfellow2014generative} and the variational autoencoder \cite{kingma2013auto}. Rather than directly generating deterministic images, the proposed InjectionGAN is designed to diversify the generation, i.e., the large variance among the generated image in all the target (even the original) domains. As demonstrated in Figure \ref{overview}, we augment the input by introducing additional information combining both unspecific factors of variation and specific ones of the target domain. The unspecific factors encapsulate the ambiguous aspects of the output learned from the whole training dataset. We inject the unspecific factors into the inputs for variation in the outputs. In other words, the proposed model implicitly injects additional information with the operations in the latent space. The specific ones specify the target domain of the output. We use a label (e.g., binary or one-hot vector) to represent the domain information, and the latent codes (e.g., random Gaussian) to represent the unspecific variation. During training, we randomly select a label indicating the target domain and draw the latent codes from a prior explicit distribution. Consequently, the many-to-many mapping is learned within one unified framework \cite{YU201781}. 

The main contributions and features of the proposed model include:
\begin{itemize} 
	\item A novel deep generative model is proposed for image-to-image translation. This model learns a many-to-many mapping function among multiple domains with unpaired training data.

	\item The proposed model unifies the generative adversarial network and variational autoencoder to explore the latent space, which is indicated by domain-specific features and unspecific random variations. 
	
	\item A novel neural network structure is developed to combine the input images with latent variables. The input of the model is a combination of the observed image, domain-specific features, and unspecific variations. Within one unified framework, the trained model generates diverse samples in multiple domains.
	
	\item The proposed model is qualitatively and quantitatively evaluated on multiple datasets with respect to style transfer and  the facial attribute transfer tasks. Its diverse generations with high quality reflect a superior performance over baseline models.
\end{itemize}

We extensively evaluate the proposed model trained on edge $\leftrightarrow$ photo, male $\leftrightarrow$ female and face $\leftrightarrow$ Emoji datasets. The experimental results demonstrate that the proposed InjectionGAN generates images with superior diversity and quality compared with the state-of-the-art methods, such as CycleGAN \cite{CycleGAN2017}, BicyleGAN \cite{zhu2017toward} and StarGAN \cite{choi2017stargan}. 

The remainder of this paper is organized as follows. We briefly review the related work and provide some background in Section \ref{background}. In Section \ref{InjectionGAN}, our proposed approach is elaborated in details. Then Section \ref{experiment} demonstrates the experimental results and analyses. Finally, this paper is concluded in Section \ref{conclusion}.

\section{Related work}\label{background}

{\bf Deep Generative Networks.} Deep generative models have shown great potential in dealing with various computer vision tasks, such as image generation \cite{GaitG,coopnets,dumoulin2016adversarially}, image translation \cite{zhu2017toward,larsen2016autoencoding} and face image synthesis \cite{tran2017DRGAN,zhang2017age}. Two types of deep generative models are distinguished among recent related researches. The first one is the generative adversarial network (GAN) \cite{goodfellow2014generative}, which takes two components to seek a Nash equilibrium within the adversarial training scheme. It minimizes the discrepancy between the distributions of generated images and the true data through penalizing the ratio of these two distributions, and thus encourages the generations to be realistic. WGAN \cite{arjovsky2017wasserstein} utilize the Wasserstein distance as the metric of the discrepancy, and it shows impressive improvement in terms of generation quality and training stability. The second one is the variational autoencoder (VAE) \cite{kingma2013auto,Yuencoder}. It learns to match an inferred posterior distribution to one explicit prior by optimizing a variational lower bound. One extension is the adversarial autoencoder \cite{makhzani2015adversarial}, which matches an integral posterior distribution to the prior through adversarial learning. WAE \cite{tolstikhin2017wasserstein} implements the similar idea by optimizing the Wasserstein distance between the latent codes distribution and the prior. The second type of approaches solve the problem of inference within the GAN framework at the cost of generation quality. We also take the advantage of GANs to guarantee the visual quality of generations.

{\bf Conditional GAN for Image-to-Image Translation.} The problem of Image-to-image translation is to change the style or texture of the image while maintain other parts largely unchanged. Multiple works \cite{kim2017learning,bousmalis2016unsupervised} employ GAN-based architectures to perform image-to-image translation between two paired/unpaired sets of image collections. 
They belong to the category of conditional GANs (cGANs), since the generated images are conditioned on the input images. 
CycleGAN \cite{CycleGAN2017} employs the cycle consistency loss to constrain the dual domain mapping and thus to train the model in an unsupervised manner; similar ideas were proposed in \cite{kim2017learning} and \cite{yi2017dualgan}. The study \cite{taigman2016unsupervised} sought to add constraint on the latent codes; \cite{liu2017unsupervised} explored the latent space by integrating the VAE and GAN. These image-to-image translation methods aim to change the image style across two different domains. The quality has been attained at the expense of diversity, as the generator learns to largely ignore unnecessary variation when conditioned on a relevant context. To enhance the diversity, the BicycleGAN \cite{zhu2017toward} succeeded in generating diverse samples in the target domain after a supervised training on paired dataset. The recent MUNIT \cite{huang2018munit} was trained in an unsupervised manner. It generates diverse samples in both the source and the target domains. However, this model is inefficient in the field of multi-domain translation as it needs to learn a separate mapping for each pair of domains. In contrast, we attempt to generate diverse samples across multiple domains within one unified model, while ensure a relatively high visual quality of the generations. 

{\bf Latent Space Exploring with GAN and VAE.}
Representation learning refers to the task of learning a representation of the data that can be easily
exploited \cite{8214214,xu2019}. Deep probabilistic models parameterize the learned representation by a neural network. \cite{wang2016generative} proposed an interpretable model to learn two types of representations: content and style of the images. An encoder-decoder structure was employed in \cite{tran2017DRGAN} to learn both the generative and discriminative representations with strong supervision. In \cite{xu2019}, the authors show that standard deep architectures can adversarially approximate to the latent space and explicitly represent factors of variation for image generation. \cite{zhu2017toward} proposed a similar framework to introduce variations to the generated images, but this approach can just learn one direction mapping with paired training dataset. 

In summary, these existing frameworks are capable of learning the relations under strong supervision. However, limited scalability has been shown in generating diverse outputs in both domains. Our approach attempts to integrate two generative models in one unified framework. It injects additional information into the latent space for diversity, and is adversarially trained to learn a many-to-many mapping between multiple domains. 

\section{Methods}\label{InjectionGAN}
\subsection{Formulation}
\begin{figure*}[t]
	
	\begin{minipage}[b]{1.0\linewidth}
		\centering
		
		\begin{minipage}[b]{1.0\linewidth}
				\centering
			{\includegraphics[width=13cm]{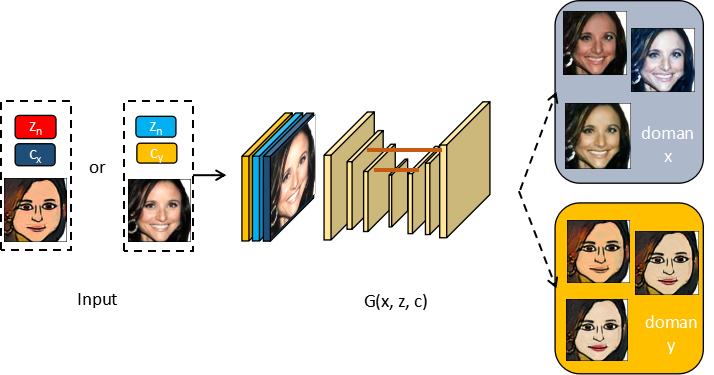}}
		\end{minipage}
	\end{minipage}
	%
	%
	\caption{The injected information consists of one specific target domain label $c$ and one unspecific variance of a latent code $z$. With the injected information, our model is able to generate samples with variation in both domains. }
	\label{Method}
\end{figure*}

In this work, we aim at learning a many-to-many mapping in a unified framework. Let us denote the domain label as $c$ and the latent random variation as $z$, which is added to avoid generating a deterministic output in the target domain. As shown in Figure \ref{Method}, The input of the generator $G$ is a combination of the observed image, the domain label $c$ and the injected random information $z$. Given $\{x_i\}_i^N$ the observed training data in the original domain and $\{y_i\}_i^N$ the observed training data in the target domain, our model contains four mappings: $G: (x,c_y,z_1) \to y'$, $G:(x,c_x,z_2) \to x'$, $G:(y,c_x,z_3) \to x'$ and $G:(y,c_y,z_4) \to y'$. These four mappings are learned simultaneously in one generator $G$, but distinguished with the input image and the target domain label. In the following section, we denote $G(x,z,c) \rightarrow y$ as the mappings without loss of generality.

At the training stage, an encoder is introduced to learn the low-dimensional latent codes $z$, which will be injected to another input for unspecific variants. We randomly generate the target domain label $c$, and then learn a variational mapping $E: (x) \rightarrow z$ to the output. To enable stochastic sampling, we desire the latent code vector $z$ to be drawn from some prior distributions $p(z)$, and we utilize a standard Gaussian distribution $\mathcal{N}(0, I)$ in this work. In addition, we introduce one adversarial discriminator $D$, aiming to distinguish between the input images ${x}$ and the generated images ${G(x,z,c)}$. Figure \ref{Method} illustrates the training process of our proposed approach. Note that the employed discriminator is coupled with an classifier, which enables the generator to produce diverse and realistic results in the domain denoted by the domain label $c$. 

\subsection{ Generative Adversarial Network}
The original GAN \cite{goodfellow2014generative} has achieved great success in producing realistic natural images \cite{radford2015DCGAN} by establishing a min-max game between the generator and the discriminator. As the discriminator learns to distinguish the increasingly accurate fake images from the real ones, the generator is regularized to yield more realistic images. The training objective is written as 
\begin{align}
\begin{aligned}
\mathcal{L}_2(X_s,X_t,&G, D) = \mathbb{E}_{x\sim X}[log D(x)] +\mathbb{E}_{z\sim p(z)}[log(1-D(G(z)))]
\end{aligned}
\end{align}
where $G$ generates image $G(z)$ conditioned on the random noise $z$, and $D$ attempts to distinguish between the real $x\sim X$ and the fake images $G(z)$.

\subsection{ GAN with Domain Classification}
The original GAN is able to tell if the image is belong to one distribution. In case of multi-domain image-to-image translation, it requires to train the generator-discriminator pairs for each domain. To this end, we extent the GAN framework for multi-domain translations. We adopt an adversarial loss conditioned on the domain labels $c \sim C$ and the encoded latent codes $z\sim E(x)$, which is written as
\begin{align}
\begin{aligned}
\mathcal{L}_{adv} = \mathbb{E}_{y\sim Y}[log D(y)] + \mathbb{E}_{x\sim X,z\sim p(z),c\sim C}[-log D(G(x,z,c))]
\end{aligned}
\end{align}
where $G$ generates an image $G(x, z, c)$ conditioned on the input image $x$, the target domain label $c$ and the random codes $z$. We attempt to translate the input images to the target domains indicated by $c$. As the target domain label $c$ is selected randomly, our model gradually learns to translate images across each domain. Moreover, adding the random code $z$ introduces variations into the model.

In addition to distinguish the real images from the fake ones, we extend the discriminator by coupling an auxiliary domain classifier. As a minimal supervision, we assume that the input image and the original domain label pairs $(x, c_0)$ are available from the training dataset. Our goal is to generate multiple images $y$ in the target domain $c$, and they are different from each other. These varieties derive from the injected $z$. According to this argument, we add the auxiliary classifier on top of $D$ and impose the domain classification loss 
\begin{align}
\begin{aligned}
\mathcal{L}_{cls}^r = \mathbb{E}_{x,c_0}[-logD_{cls}(c_0|x)]
\end{aligned}
\end{align}
\begin{align}
\begin{aligned}
\mathcal{L}_{cls}^g = \mathbb{E}_{x\sim X,z\sim p(z),c\sim C}[-logD_{cls}(c|G(x,z,c))]
\end{aligned}
\end{align}
where the term $D_{cls}(c_0|x)$ represents a probability distribution over domain labels computed by $D$. By minimizing this objective, $D$ is trained to enforce the generated images conditioned on the domain label. Note that the target domain label $c$ could be either the original domain label $c_0$ or a different one. As a result, the generator $G$ produces diverse images in each domain through manually assigning the domain label $c$.

\subsection{Conditional VAE-GAN with Latent Consistency}
Injecting the random codes $z$, which contain the colors and textures information, introduces variations into the model. To learn these latent semantic representations, we resort to variational autoencoder. In VAE model, generating samples can be seen as the regrouping of the latent codes $z$ given the input image x. The joint distribution over the pair $(x, z)$ is given by $p(x, z) = p(x | z)p(z)$, where $p(z)$ is the prior distribution over the latent variables $z$ and $p(x | z)$ is the conditional likelihood function of the input image $x$. Let $\eta$ be a random vector with a multi-variant Gaussian distribution: $\eta \sim  \mathcal{N}(\eta|0, I)$. The sampling operation of $z \sim q(z|x)$ can be implemented via $z$ = $E(x) + \eta$, using the re-parameterization trick, allowing direct back-propagation \cite{kingma2013auto}. Thus, we have $z \sim E(x) = q(z|x)$ and $x \sim G(z) = p(x|z)$. The networks are trained by minimizing the upper-bound on the expected negative log-likelihood of $x$, which is given by
\begin{align}
\label{eq4}
\begin{split}
\mathcal{L}_1(X,E,D) &= \mathbb{E}q_(z|x)[-logp(x|z)] + KL(q_(z|x)||p(z) ) 
\end{split}
\end{align}
where the first term in Eqn. (\ref{eq4}) corresponds to the reconstruction error, and the second term is a regularization term that encourages the posterior distribution to approximate the prior distributions. Next, we describe the two terms in details. 

Injecting random variations into the generator, on one hand, diversifies the generations; on the other hand, makes the system unstable. Since the latent codes are randomly drawn, we minimize the cyclic loss to match the reconstructed image to the input image 
\begin{align}
\label{eqx}
\begin{split}
\mathcal{L}^{image}_{1} &= \mathbb{E}_{x\sim X, y \sim G(x,z,c), z \in E(x)}||x-G(y,z,c_0)||_1
\end{split}
\end{align}

The encoded latent codes distribution from the $E$, in theory, should match a random Gaussian, which will enable sampling at the inference time, when the data distribution is intractable. It is important to enforce the latent codes to learn the information for variations and it is also equally important to force the encoded latent codes distribution approximate the prior distribution $p(z)$. To this end, we optimize the model by penalizing the $KL$ divergence between the encoded codes distribution and the prior
\begin{align}
\label{eq2}
\begin{split}
\mathcal{L}_{KL} &= \mathbb{E}_{x\sim X,z\sim N(0,I)}[D_{KL}(E(x)|| p(z))]
\end{split}
\end{align}

In our work, we directly encode the generated image as $E(y) = E(G(x,c,z)$, and enforce the latent codes $z'=E(y)$ to match the random sample $z$ from the prior $p(z)$. Consequently, the latent codes $z$ are encrypted with the unspecific variations (colors or textures), and the generator $G$ taking the information containing latent codes $z$ and the domain label $c$ yields diverse outputs in the target domain. Specifically, we start from a randomly drawn latent codes $z$ and attempt to recover it by requiring $z'= z$, which we denote as the latent code consistency. 
\begin{align}
\label{eq1}
\begin{split}
\mathcal{L}^{latent}_{1} &= \mathbb{E}_{x\sim X, z \sim p(z), c\sim C}||z-E(G(x,z,c))||_1
\end{split}
\end{align}

The above objective function shares the key idea of latent consistency, where the encoded codes should be consistent with the randomly drawn latent codes. In this work, we employ the same operation to learn the latent codes in all domains.


%
%
%

\subsection{ Overall Training Objective}
The trained model translates the input images to different domains and yields realistic images with variations introduced by the latent code $z$. During training, we combine the domain classification objectives and the conditional VAE-GAN objective, and minimize a weighted sum of the above losses:
\begin{align}
\label{eq3}
\begin{split}
\mathcal{L}_{D} &= -\mathcal{L}_{adv} + \lambda \mathcal{L}_{cls}^{r} 
\end{split}
\end{align}
\begin{align}
\label{eq5}
\begin{split}
\mathcal{L}_{G,E} = -\mathcal{L}_{adv} + \lambda_{cls} \mathcal{L}_{cls}^{g}  + \lambda_{image} \mathcal{L}_{1}^{image}+  \lambda_{latent} \mathcal{L}_{1}^{latent} + \lambda_{KL} \mathcal{L}_{KL}
\end{split}
\end{align}
where $\lambda_{cls}$, $\lambda_{image}$, $\lambda_{KL}$ and $\lambda_{latent}$ are hyper-parameters that control the
relative importance of domain classification and information injection, respectively, compared to the adversarial loss.
We use $\lambda_{image} = 1$, $\lambda_{latent} = 10$, $\lambda_{KL}=0.5$ and $\lambda_{cls} = 2.5$ in all of our experiments.




\section{Experimental Evaluation}\label{experiment}
Our model aims at diversifying the translated images among different domains. In this section, we discuss the implementation in details, and compare the performance with the state-of-the-art methods on multiple datasets. We first quantitatively evaluate the performance on two datasets in terms of diversity and visual quality of the generations. Then we analyze the effects of the $\mathcal{L}_1^{latent}$ loss and the neural network architecture on the diversity. We finally demonstrate the generality of our algorithm on applications where paired data does not exist. 

\subsection{Implementation}
{\bf Improved GAN Training.} GAN models are notoriously hard to train. To stabilize the training process for generations with higher quality, we take the advantage of Wasserstein GAN with gradient penalty (WGAN-GP) \cite{gulrajani2017improved}, which is defined as
\begin{align}
\begin{aligned}
\mathcal{L}_{adv} &= \mathbb{E}_{y\sim Y}[D(y)] + \mathbb{E}_{x\sim X,z\sim p(z),c\sim C}[-D(G(x,z,c))]\\&+ 
\lambda_{gp}\mathbb{E}_{\hat{x}}[(||\triangledown_{\hat{x}}D(\hat{x})||_2-1)^2]
\end{aligned}
\end{align}
where $\hat{x}$ is the linear interpolation between a real image and a synthesized one. In our experiments, We set $\lambda_{gp} = 5$ for all tasks.

{\bf Network Architecture.} The generator $G$ is based on the U-Net \cite{ronneberger2015u}, which comprises of an encoder-decoder architecture. This structure is able to explore the latent space while maintain the spatial correspondence between the input and output with skip connections. The encoder and decoder are composed of several convolutional layers with stride size of two for downsampling and upsampling, respectively. Each convolutional layer of the decoder shares the component of the corresponding part of the encoder with skip connection. For the discriminator $D$, we leverage on patch discriminator \cite{pix2pix2016,CycleGAN2017}, which discriminates on whether the image patches are real or fake instead of the whole image. We add one additional convolutional layer after the second latest convolutional layer of the discriminator as the domain classifier. The encoder $E$ comprises of a few convolutional and downsampling layers, which ensure the exploration of the latent space.

{\bf Datasets.} To demonstrate the plausibility of the proposed method, we evaluate the model qualitatively and quantitatively on two existing open datasets, and one collected dataset. All the tasks focus on diversifying the generations while maintaining high image quality. The learned many-to-many mappings include: edge $\leftrightarrow$ photo, male $\leftrightarrow$ female, and face $\leftrightarrow$ Emoji \cite{taigman2016unsupervised}. The datasets are organized as 

\begin{itemize} 
	\item Edge $\leftrightarrow$ Photo. The dataset is downloaded from \cite{yu2014fine} and \cite{zhu2016generative}. It contains paired images where the ground truth of input-output pairs are available. Here, we train the model without using the paired information, instead, we utilize the ground truth for comparison.
	\item Male $\leftrightarrow$ Female. The CelebA \cite{liu2015deep} dataset is for attribute-based translation of face images. In our experiment, each image contains a male face or female face. We perform the many-to-many translation by learning the mapping between the male and female attributes.
	\item  Face $\leftrightarrow$ Emoji, inspired by \cite{taigman2016unsupervised}. Our model is trained to learn the mapping on the data from two different domains. One domain is composed of real-world frontal-face images from the CelebA dataset. As a preprocessing step, we align the faces based on the location of eyes and mouth, and remove the background. The other domain contains the Emoji and we take the same preprocessing step to align the Emoji.  For each domain, the corresponding training image collection consists of 1,200 images.		
\end{itemize}

\begin{table*}[t]
	\renewcommand{\tabcolsep}{1mm}
	\centering
	\caption{Comparison with baseline methods with respect to the training datasets and the learned mapping function. The row starts with ``paired data" points out the requirement on the dataset. the one-to-one represents the one domain to one domain translation, while many-to-many represents the many domains to many domains translation. }
	\label{methods}
	\begin{tabular}{|c|cccccc|}
		\hline
		\multirow{ 2}{*}{Method} &\multirow{ 2}{*}{Ours}& \multirow{ 2}{*}{StarGAN}&\multirow{ 2}{*}{CycleGAN}&\multirow{ 2}{*}{BicycleGAN}&\multirow{ 2}{*}{pix2pix}&\multirow{ 2}{*}{pix2pix+} \\
		&&&&&& \\
		\hline
		\hline
		Paired data &  &&&\checkmark&\checkmark&\checkmark\\
		One-to-one  && &\checkmark&&\checkmark&\\
		One-to-many &&\checkmark&&\checkmark&&\checkmark\\
		Many-to-many &\checkmark &&&&&  \\
		\hline
	\end{tabular}
\end{table*}
{\bf Training details.} We train the models using Adam \cite{kingma2014adam} with $\beta_1 = 0.5$ and $\beta_2 = 0.999$, and we train all models for 100 epochs with a learning rate of 0.0001 and apply the same decaying strategy over the next 100 epochs. Following the training scheme proposed in \cite{arjovsky2017wasserstein,gulrajani2017improved}, we update the generator five times then update the discriminator one time. We train all the models on $128 \times 128$ images and the batch size is set to 16. For data augmentation, we first resize the image to $138\times138$, then random crop the image to $128\times128$, and flip the images horizontally with a probability of 0.5. Training is performed on a single NVIDIA Tesla K80 GPU.

{\bf Compared methods.} We take multiple state-of-the-art models as baselines. Table \ref{methods} illustrates the characteristics of each approach. BicyleGAN \cite{zhu2017toward} diversifies the generations by exploring latent variables, but it relies on paired training images to learn an one-directional mapping, which is an extension of Pixel2pix \cite{pix2pix2016} and Pixel2pix+noise. The CycleGAN \cite{CycleGAN2017} learns bidirectional mapping by employing unpaired training images. This model translates the input images between two domains at one training, but it needs multiple steps to train models for multiple domain translation. A recent work of StarGAN \cite{choi2017stargan} addresses this limitation by learning a unified structure that maps input images to multi-domains. However, it dose not explore the latent space and fail to generate diverse samples in each domain. While the proposed model is a unified architecture to diversify generations in multiple domains (i.e., combining the latent variables and the domain label as a vector). It does not require pairwise training data for learning multi-domain mapping. 

{\bf Evaluation metrics.}
To quantitatively evaluate the performance in terms of the generation diversity, we report the Learned Perceptual Image Patch Similarity (LPIPS) metric proposed in \cite{zhang2018unreasonable} as the diversity score. This metric is a perceptual distance to assess diversity of the outputs. For each method, we compute the average distance between $1,900$ randomly generated output images. 

In order to quantitatively assess the quality of the synthesized images, we employ the Frechet Inception Distance (FID) introduced in \cite{heusel2017gans}, which is a measure of similarity between the images of two datasets. It is shown to correlate well with human judgment of realism and faithfulness of the generations. We report the results based on $10,000$ random generated samples.

\subsection{Quantitative Comparison against Baselines}
In this subsection, we quantitatively compare the performances against baseline methods on the Edge $\leftrightarrow$ photo and Female $\leftrightarrow$ Male datasets. We obtain the generation $G(x,c,z)$ with the random codes $z$, the domain label $c$ and the input image $x$. The codes $z$ are randomly drawn from Gaussian distribution, and the target domain label $c$ is assigned to indicate either the original or a different domain. By regrouping the random codes and target domain label at each time, the model generates different samples in the target domain.  

\subsubsection{Edge $\leftrightarrow$ Photo}
\begin{figure*}[t!]
	
	\begin{subfigure}[t]{1.0\textwidth}
		\centering
		{	\centerline{\includegraphics[width=12cm]{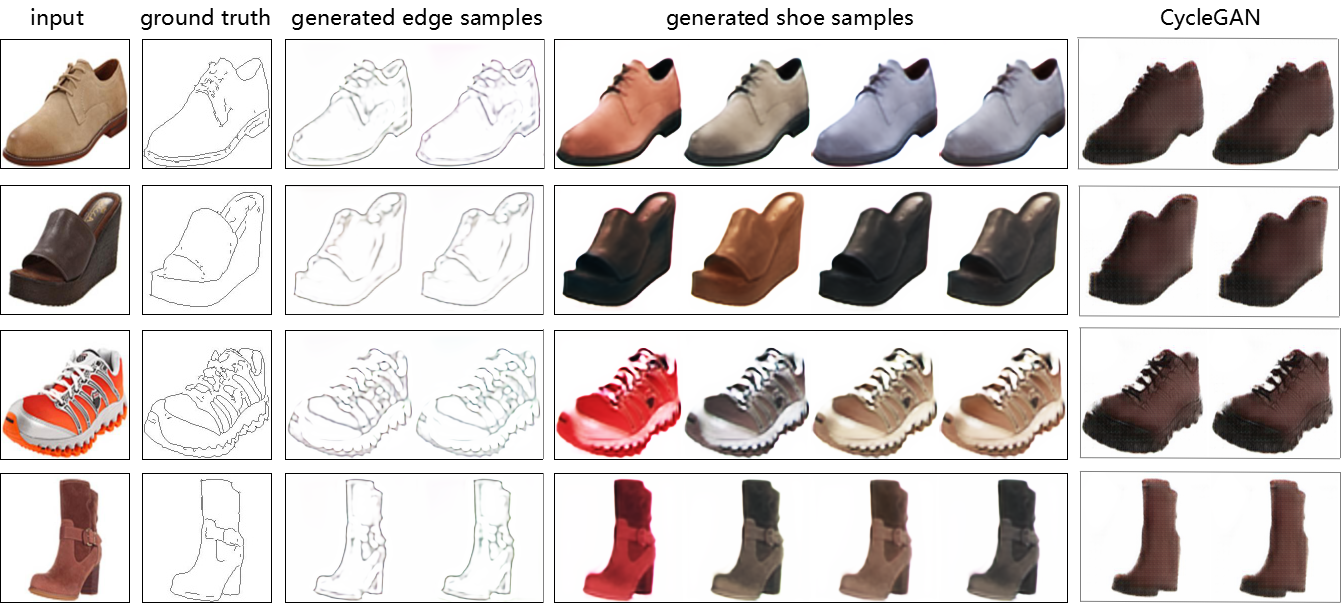}}	}
		
		\caption{\label{f_edge2shoe_1}Given edge images as inputs, our model generates samples in both domains. It is evident that the translated generations are able to capture different colors, while the samples generated by the CycleGAN do not have such variation.}		
	\end{subfigure}

	\begin{subfigure}[b]{1.0\linewidth}
		\centering
		{\centerline{\includegraphics[width=12cm]{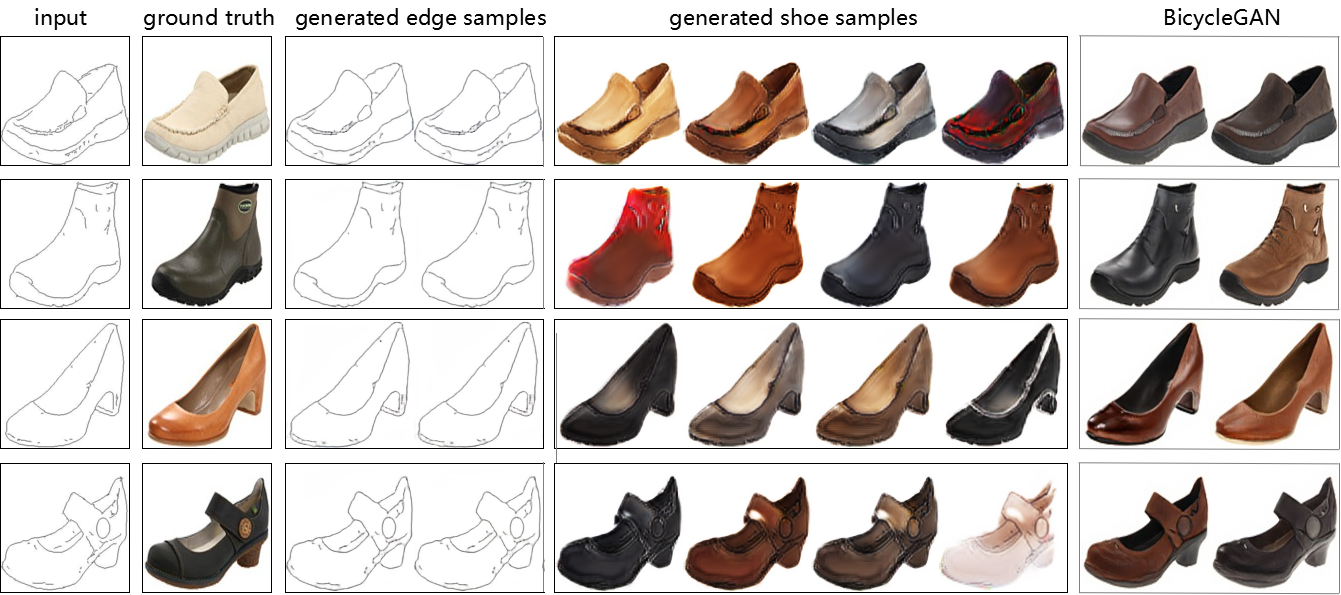}} 	}
		\caption{\label{f_edge2shoe_2}The proposed model takes the edge images as input and achieves diversity in the photo domain. Note that the BicycleGAN is trained with paired dataset and can only learn an one-direction mapping (Edge $\to$ Photo).}
	\end{subfigure}
	\caption{Demonstration of the diverse generations from the 'edge $\leftrightarrow$ photo' task.}
	\label{f_edge2shoe}
\end{figure*}

We learn a unified bi-direction mapping, which translate the images between four domain pairs in one framework. This model is trained without any pairing information, even though the paired images are provided in this dataset. To illuminate the learned mappings, we qualitatively show the input images and the translated images in both domains. Figure \ref{f_edge2shoe_1} presents the generations from two mappings: photo $\to$ edge and photo $\to$ photo. Figure \ref{f_edge2shoe_2} contains the generations from the other two mappings: edge $\to$ edge and edge $\to$ photo. All these four mappings are learned in one generator $G(x,c,z)$.

\begin{table}[t!]
	\renewcommand{\arraystretch}{1.2}
	\centering
	\caption{Realism vs diversity. We measure diversity using average LPIPS distance, and realism using a
		real vs. fake on the Edge $\to$ Photo task. The pix2pix+noise baseline
		produces little diversity. The hybrid
		BicycleGAN method produces results which have higher realism
		while maintaining diversity}
	\label{edge2shoe}	
	\vspace{2.0mm}
	\begin{tabular}{|p{5cm}p{2cm}p{1.5cm}|}
		\hline
		&Realism&Diversity\\
		\hline
		
		Method &FID score& LPIPS\\
		\hline
		\hline
		Random real img &1.26 &0.265\\
		\hline
		pix2pix+noise  &20.67  &  0.012\\
		pix2pix  & 14.50 &  0.009\\						
		CycleGAN  & 31.04 &  0.009\\
		MUNIT    &12.84 & 0.109 \\
		BicycleGAN &  8.39   &  0.111\\	
		InjectionGAN $w/o$ $l^{image}$ &    41.25&  0.139\\		
		InjectionGAN $w/o$ $l^{latent}$ &27.45  &  0.095\\					
		InjectionGAN $w/o$ $l_{KL}$ &23.01  &  0.090\\						
		InjectionGAN (edge$\to$photo) &  20.25 &  0.107\\
		InjectionGAN (photo$\to$photo)& 11.49  &  0.088\\										
		\hline
	\end{tabular}
	%
	%
\end{table}

{\bf Diversity.} For comparison, all the baseline models are trained to map the edge images to shoe images, and the InjectionGAN generates photo images in two directions: edge$\to$photo and photo$\to$photo. We report the LPIPS score of the translated photo images as the measurement of diversity. The LPIPS scores for each method are listed in Figure \ref{edge2shoe}. Random ground truth shoe images produce an average variation of $0.265$. The pix2pix and CycleGAN systems yield fixed point estimates in the photo domain, which is proved by the low LPIPS scores. Adding noise to the system pix2pix+noise slightly increase the score, confirming that adding noise without any encoded information does not produce large variations. The InjectionGAN models put explicit constraints on the latent space. This method produces higher diversity scores than the cGANs regardless of the quality. Since unnatural images containing meaningless variations may yield high diversity scores, we next compare the performances in terms of the visual realism of the generations.  

{\bf Realism.} The FID scores are listed in Table \ref{edge2shoe}, which reflect the visual quality and faithfulness of the synthesized images. The BicycleGAN performs the best in terms of FID score. Note that it takes paired images for training. The pix2pix+noise and CycleGAN models, translating images at the pixel level, achieve high realism score. The InjectionGAN achieves competitive performance with a little lower FID score as it learns the latent space as a structure. Comparisons between these models show that the autoencoder based models explore the latent space, which introduces uncertainty to the model. They generate diverse outputs but the FID scores of the synthesized images are relatively lower, while the cGANs, translating images in pixel level, yield images with high quality at the cost of diversity.
\subsubsection{Male $\leftrightarrow$ Female}
\begin{figure}[t!]
	\begin{minipage}[b]{1.0\linewidth}
		\centering
		\centerline{\includegraphics[width=13cm]{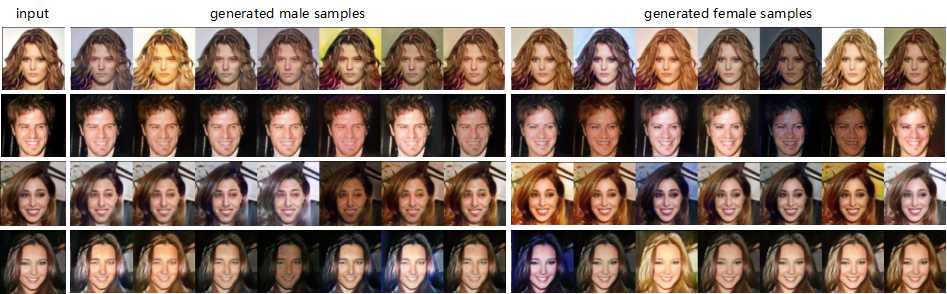}}
		
	\end{minipage}
	\caption{Giver a male or female face, we generate 7 128 x 128 faces in each domain. }
	\label{male2female}
\end{figure}

We employ our model to learn another bi-directional mapping between the male and female images. This model is trained with the same hyper parameters. Figure \ref{male2female} shows the synthesized images. It is evident that our model introduces meaningful variations in the generated faces without compromising the visual quality. The quantitative results are reported in Table \ref{table:male2female}. The results demonstrate that the proposed model generates diverse faces in both domains, which highlights the importance of injecting additional information. We then quantitatively evaluate the effects of the network architecture and the $\mathcal{L}_1^{latent}$ on diversity and quality. 

\begin{table}[t!]
		\renewcommand{\arraystretch}{1.2}
			\centering
		\captionof{table}{Comparison to Baseline methods on the male $\leftrightarrow$  female task.}
		\label{table:male2female}		
		\begin{tabular}{|p{6cm}p{1cm}p{1.2cm}|}
			\hline
			
			Method & FID& LPIPS\\
			\hline
			\hline
			Real img &1.26 &0.254\\
			\hline
			CycleGAN & 7.58 &  0.031\\	
			InjectionGAN (male $\to$ female) & 10.25   &  0.108\\
			InjectionGAN (male $\to $male)&  6.49   &  0.069\\						
			\hline
		\end{tabular}
\end{table}

	\begin{table}[t!]
		\centering
		\captionof{table}{The performance with respect to different generator architectures and weights
			of $\mathcal{L}_1^{latent}$.}
		\label{latent}		
		\begin{tabular}{|p{1.9cm}p{2.5cm}p{1cm}p{1.2cm}|}
			\hline
			Method &Weight&FID& LPIPS\\
			\hline
			\hline
			Real img &&1.26 &0.254\\
			\hline
			&$\lambda_{latent} = 0.5$  &  8.38&  0.011\\
			ResNet&$\lambda_{latent} = 5$ &  9.38&  0.013\\	
			&$\lambda_{latent} = 10$ & 10.39   &  0.027\\
			\hline
			&$\lambda_{latent} = 0.5$  & 7.12   &  0.039\\
			Unet&$\lambda_{latent} = 5$  & 10.25   &  0.108\\		
			&$\lambda_{latent} = 10$ & 34.30   &  0.214\\						
			\hline
		\end{tabular}
	\end{table}

\begin{figure*}
	\centering
	\begin{subfigure}[t]{0.43\textwidth}
		\centering
		\includegraphics[height=4.2cm]{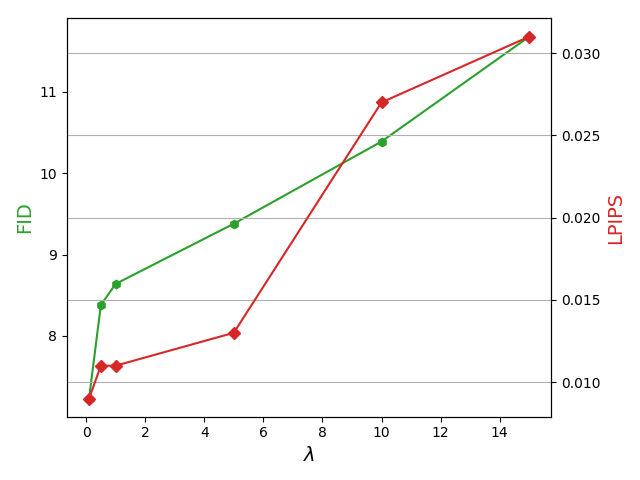}
		\caption{\label{Resnet}Resnet}
	\end{subfigure}
	\hspace{5mm}
	\begin{subfigure}[t]{0.45\textwidth}
		\centering
		\includegraphics[height=4.2cm]{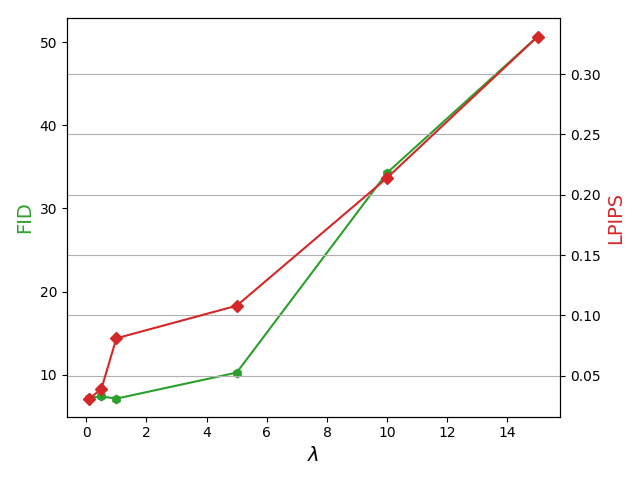}
		\caption{\label{Unet}Unet}
	\end{subfigure}
\caption{The effects of different $\lambda$ on the the performance of Resnet and Unet.}
\label{fig7}
\end{figure*}

{\bf Analysis of the loss function.}
In Table \ref{latent}, we compare against the latent loss with different weights $\lambda_{latent}$. Increasing the latent loss substantially degrades the quality of results, while enhancing the diversity. We therefore conclude that this latent loss term is critical for the diversity of our results. In the other experiments, we set $\lambda_{latent}=5$ as a trade-off between the generations' diversity and quality.

{\bf Resnet vs Unet.}
Our model learns the many-to-many mappings by employing the U-Net as the base architecture, which enables our model to generate samples $G(x,c,z)$ partially controlled by the randomly drawn latent codes $z$. By learning the latent space a structure, our model yields stochastic generations in both domains. By contrast, the cGANs utilizing the residual blocks for pixel level translation, such as Cyclegan \cite{CycleGAN2017} and StarGan \cite{choi2017stargan}, make fixed generations for each input image. We observe that the generated images using residual blocks are very close to the original inputs with a low LPIPS score, which is evident in Figure \ref{fig7}.

\begin{figure*}[t!]	
	\begin{subfigure}[b]{1.0\linewidth}
		\centering
		{	\centerline{\includegraphics[width=11cm]{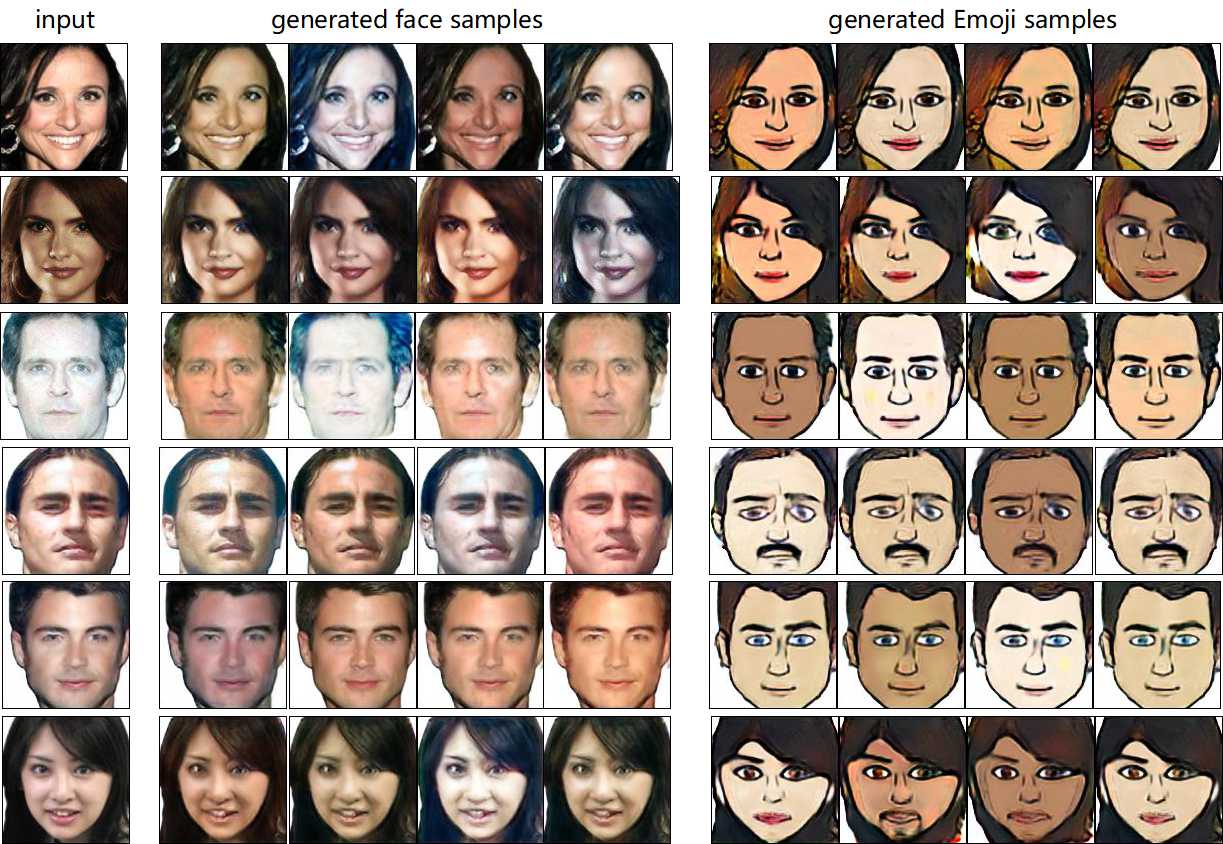}}	}
				\caption{\label{emoji_1}Example results on unpaired datasets
			of emoji $\leftrightarrow$ face.}
		
	\end{subfigure}
	\begin{subfigure}[b]{1.0\linewidth}
		\centering
		{\centerline{\includegraphics[width=11cm]{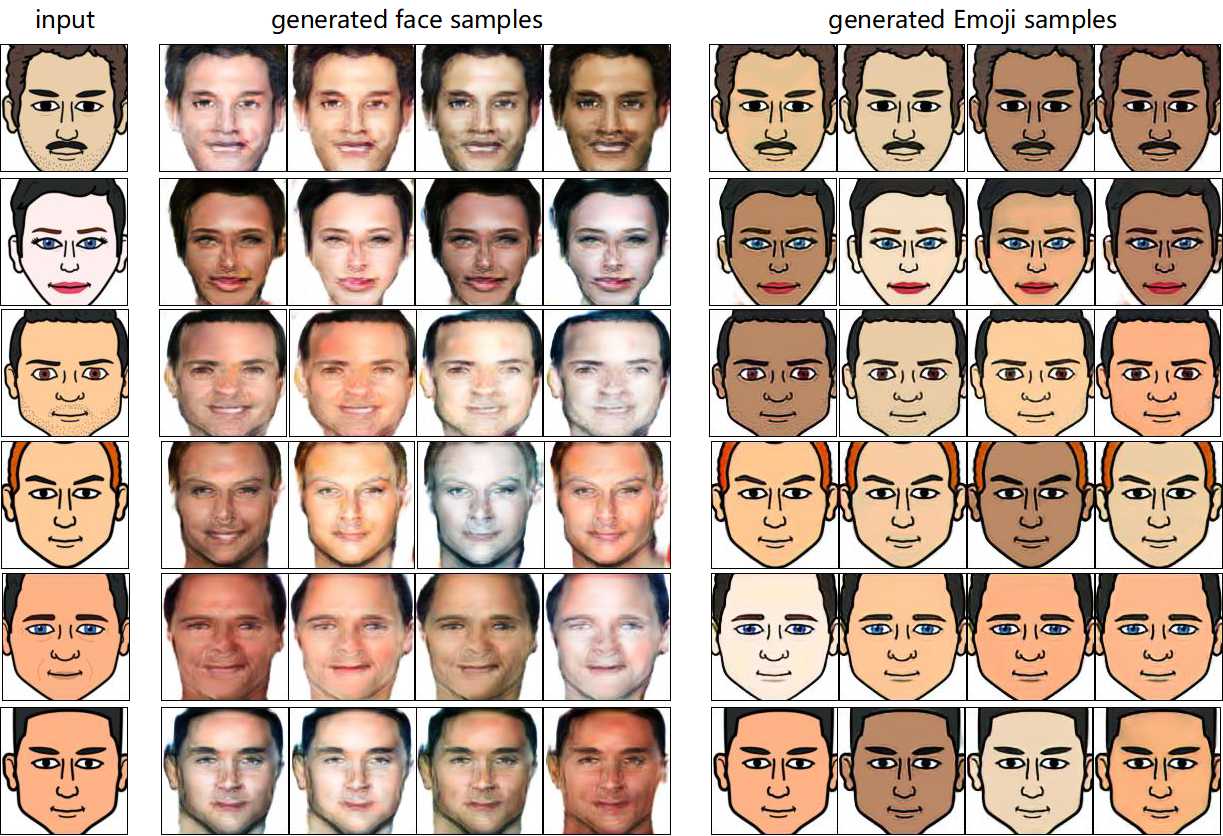}} 	}
				\caption{\label{emoji_2}Given an photo as input, we generate samples in both domains.}
	\end{subfigure}
	%
	%
	\caption{Example results on unpaired datasets
		of emoji $\leftrightarrow$ face.}
	%
\end{figure*}

\subsection{Qualitative Evaluation on the Application of Unpaired Dataset}
We qualitatively demonstrate the proposed method on three applications where paired training data dose not exist. All the generated samples are the outputs of $G(x,c,z)$ given the randomly drawn codes $z$, assigned target domain label $c$ and the input image $x$. We observe that translations with diversity are often more appealing than those generations without uncertainty.

{\bf Face $\leftrightarrow$ Emoji.} This application is inspired by \cite{taigman2016unsupervised}, which translates human face to Emoji in one direction. This model is trained with an additional constraint on the consistency of latent codes so as to reserve the facial characteristics. However, the generated images are fixed samples and not exactly faithful to the input images. We learn the proposed model as a bi-directional mapping. In one direction, it can be employed to translate the photos of human face to Emoji and translate the generated Emoji back to photos. In another direction, it is able to start with the Emoji as input and generate diverse images in both domains. In Figure \ref{emoji_1} and Figure \ref{emoji_2}, we list eight randomly translated samples for each input image. Taking the Emoji as input, our model faithfully generates diverse translated human face images along the direction Emoji $\to$ photo. In addition, it produces Emoji with different colors in the original domain with the mapping Emoji $\to$ Emoji. Conditioned on the photo as input, our model can also introduce diverse generations in both domains.  

\begin{figure*}[t]
	
	\begin{minipage}[b]{1.0\linewidth}
		\centering
		\centerline{\includegraphics[width=12cm]{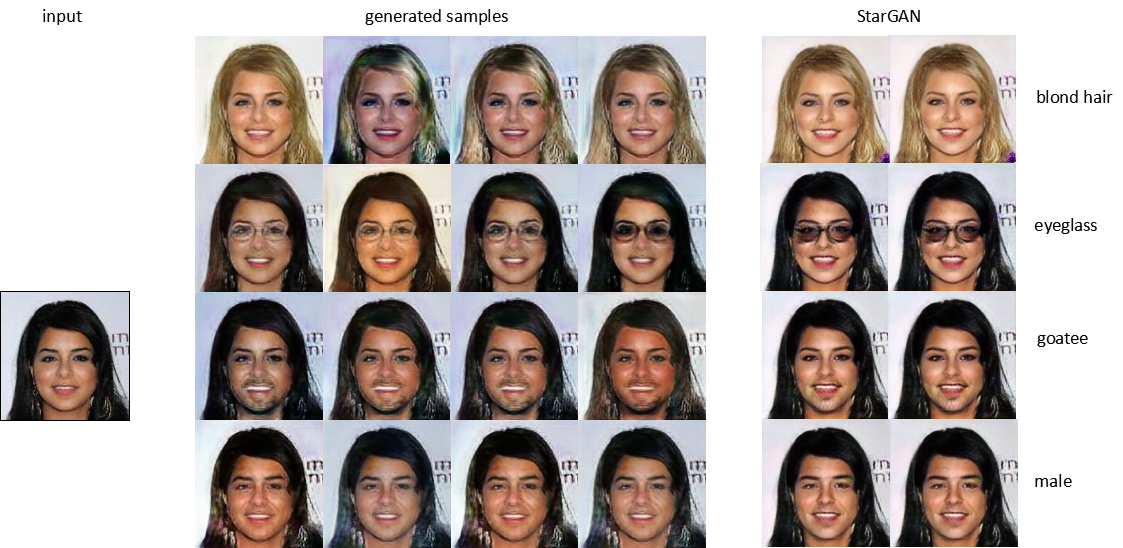}}
	\end{minipage}
	\caption{Conditional generation given attributes. We sample a set of attributes from the data distribution and generate 4 faces for each attributes. }
	\label{star5}
	%
	%
\end{figure*}
{\bf Facial attribute transfer.}
Facial attribute transfer is one of the challenging multi-domain transformation task as it requires the model to map the input to several domains with one unified framework. The pixel2pixel and CycleGAN are able to learn a one to one mapping at one training step, and need to train several times for each pair of domains within the multi-domain translation task. We evaluate the proposed method on the facial attribute transfer task. As a baseline method, we train the cross-domain models StarGAN \cite{choi2017stargan}, which learns a one-to-many mapping function in one framework. Conditioned on the inputs, the trained model generates images with the selected attributes.

\begin{figure*}[ht!]	
	\begin{subfigure}[b]{1.0\linewidth}
		\centering
		{	\centerline{\includegraphics[width=8cm]{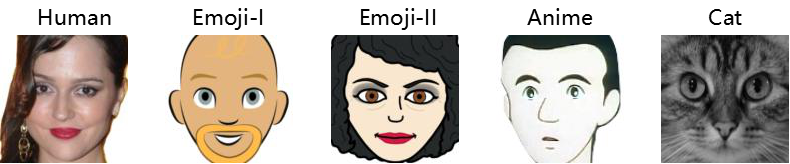}}	}
		\caption{\label{face_0}The face images from each domain of this dataset.}
		
	\end{subfigure}
	\begin{subfigure}[b]{1.0\linewidth}
		\centering
		{\centerline{\includegraphics[width=12cm]{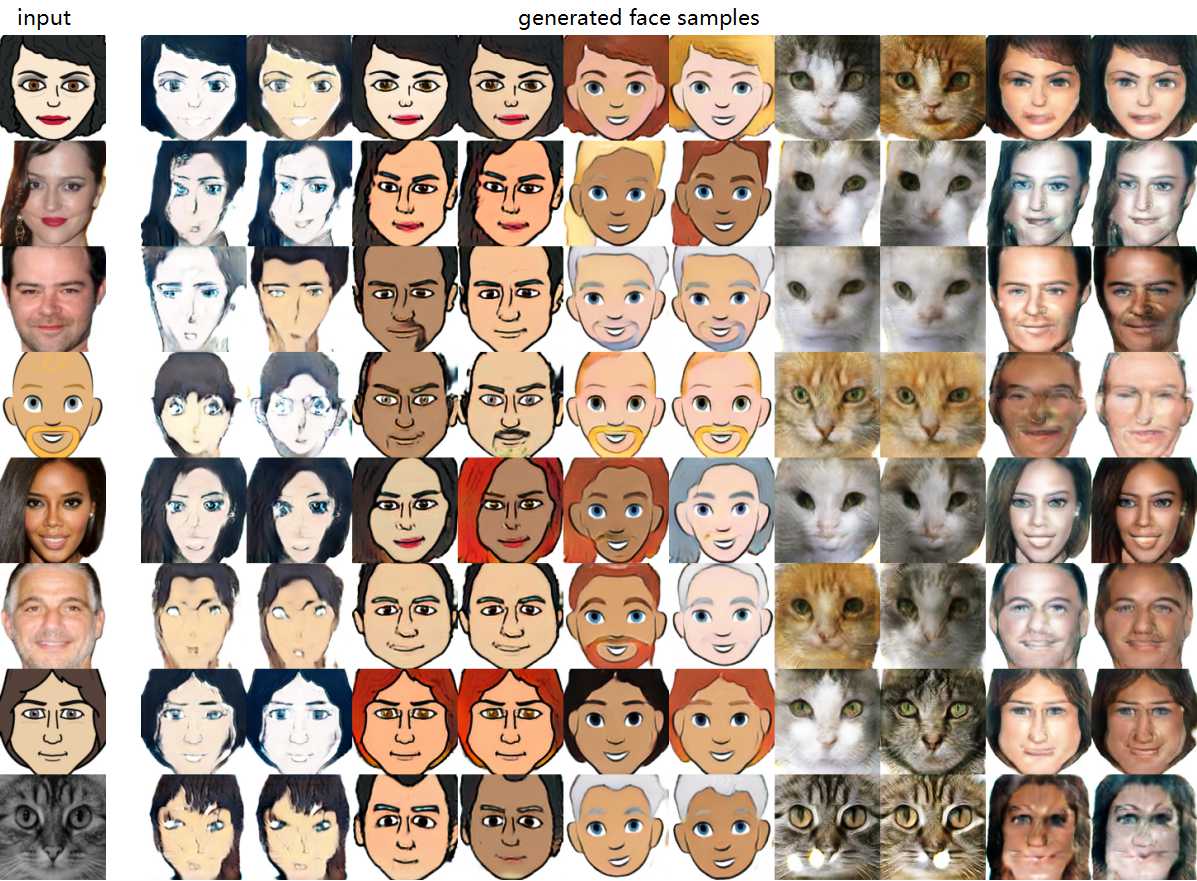}} 	}
		\caption{\label{face}Given an photo as input, we generate samples in all domains.}
	\end{subfigure}
	%
	%
	\caption{Diverse generation for face style translation task.}
	%
\end{figure*}

In Figure \ref{star5}, we visualize the results where we translate several images that do not have blond hair, male, goatee, and eyeglass to the corresponding images with each of the individual attributes. We notice that our method not only provides a high visual quality of translation results in each target domain, but also outperforms the StarGAN in terms of diversity. We observe that the StarGAN successfully translates the input images into each domain but without any variations within each target domain. In other words, rather than training a model to perform a fixed translation (e.g., input $\to$ blond hair), which is prone to over-fitting, we train our model to flexibly translate images conditioned on the target domain label and randomly drawn latent codes. This allows our model to learn reliable features universally applicable to multiple domains of images with different facial attribute values.

{\bf Face style translation.} In this task, our model is able to change the style of input images based on the injected random codes z and domain label c across five domains. Figure \ref{face} presents the synthesized results from the model trained on the dataset, which contains photos from five different sources, namely the natural human face, two Emoji-style faces , Anime face, and cat face. Different to the existing cGANs, which needs to learn a bi-directional mapping for each pair of domains, our model learns a universal mapping within one unified generator. Moreover, injecting latent codes encourages to change the color composition for diversity. For example, given one input image and the target domain label, the generator often maps the input image to the target domain with different colors, because such a many-to-many mapping is equally valid under the domain classification loss and the adversary loss.  

\section{Conclusion}\label{conclusion}
To diversify the generation in the image-to-image translation, we have proposed the InjectionGAN, which learns a many-to-many mapping in one unified framework. This model learns the additional information, such as the colors and textures, into the latent codes and injects the information into another image for diversity. Benefiting from the domain label, we can manually select the target domain so as to inject these variations into any domain. The experimental results illustrate that injecting additional information into the input image is a promising approach for many-to-many image translation tasks which involve color and texture changes. The proposed method succeeds in diverse generations and outperforms the state-of-the-art approaches. When combining with the domain label, the model also induces changes in the original domain, which enables a great potential for a wide variety of applications. As a follow-up study, it is appealing to learn a mapping across multiple domains with large domain gap, such as the objects' shape (cat $\leftrightarrow$ dog). However, both the Unet and Resnet are not effective to learn this mapping. One possible way is to increase the receptive field in the discriminator. This task is very challenging in terms of diversity and faithful reconstruction. We hope to learn a unified model to enable effective and efficient translation across multiple domains with large domain gap in the future.

\clearpage
\bibliographystyle{splncs04}
\bibliography{InjectionGAN_arXiv}

\begin{thebibliography}{10}
\providecommand{\url}[1]{\texttt{#1}}
\providecommand{\urlprefix}{URL }
\providecommand{\doi}[1]{https://doi.org/#1}

\bibitem{arjovsky2017wasserstein}
Arjovsky, M., Chintala, S., Bottou, L.: Wasserstein generative adversarial
  networks. International Conference on Machine Learning  (2017)

\bibitem{bousmalis2016unsupervised}
Bousmalis, K., Silberman, N., Dohan, D., Erhan, D., Krishnan, D.: Unsupervised
  pixel-level domain adaptation with generative adversarial networks. The IEEE
  Conference on Computer Vision and Pattern Recognition  (2017)

\bibitem{choi2017stargan}
Choi, Y., Choi, M., Kim, M., Ha, J.W., Kim, S., Choo, J.: Stargan: Unified
  generative adversarial networks for multi-domain image-to-image translation.
  Proceedings of the IEEE Conference on Computer Vision and Pattern Recognition
   (2018)

\bibitem{dumoulin2016adversarially}
Dumoulin, V., Belghazi, I., Poole, B., Mastropietro, O., Lamb, A., Arjovsky,
  M., Courville, A.: Adversarially learned inference. International Conference
  on Learning Representations  (2017)

\bibitem{gatys2016image}
Gatys, L.A., Ecker, A.S., Bethge, M.: Image style transfer using convolutional
  neural networks. Conference on Computer Vision and Pattern Recognition
  (2016)

\bibitem{goodfellow2014generative}
Goodfellow, I., Pouget-Abadie, J., Mirza, M., Xu, B., Warde-Farley, D., Ozair,
  S., Courville, A., Bengio, Y.: Generative adversarial nets. Advances in
  neural information processing systems  (2014)

\bibitem{gulrajani2017improved}
Gulrajani, I., Ahmed, F., Arjovsky, M., Dumoulin, V., Courville, A.C.: Improved
  training of wasserstein gans. Advances in Neural Information Processing
  Systems pp. 5769--5779 (2017)

\bibitem{he2018learning}
He, L., Wang, G., Hu, Z.: Learning depth from single images with deep neural
  network embedding focal length. IEEE Transactions on Image Processing
  \textbf{27}(9),  4676--4689 (2018)

\bibitem{heusel2017gans}
Heusel, M., Ramsauer, H., Unterthiner, T., Nessler, B., Hochreiter, S.: Gans
  trained by a two time-scale update rule converge to a local nash equilibrium.
  Advances in Neural Information Processing Systems  (2017)

\bibitem{huang2018munit}
Huang, X., Liu, M.Y., Belongie, S., Kautz, J.: Multimodal unsupervised
  image-to-image translation. European Conference on Computer Vision  (2018)

\bibitem{pix2pix2016}
Isola, P., Zhu, J.Y., Zhou, T., Efros, A.A.: Image-to-image translation with
  conditional adversarial networks. IEEE Conference on Computer Vision and
  Pattern Recognition  (2017)

\bibitem{kim2017learning}
Kim, T., Cha, M., Kim, H., Lee, J., Kim, J.: Learning to discover cross-domain
  relations with generative adversarial networks. International Conference on
  Machine Learning  (2017)

\bibitem{kingma2014adam}
Kingma, D., Ba, J.: Adam: A method for stochastic optimization. International
  Conference on Learning Representations  (2015)

\bibitem{kingma2013auto}
Kingma, D.P., Welling, M.: Auto-encoding variational bayes. International
  Conference on Learning Representations  (2014)

\bibitem{larsen2016autoencoding}
Larsen, A.B.L., S{\o}nderby, S.K., Larochelle, H., Winther, O.: Autoencoding
  beyond pixels using a learned similarity metric. International Conference on
  Machine Learning  \textbf{48},  1558--1566 (2016)

\bibitem{Yuencoder}
Li, F., Qiao, H., Zhang, B.: Discriminatively boosted image clustering with
  fully convolutional auto-encoders. Pattern Recognition  (2018)

\bibitem{li2017diversified}
Li, Y., Fang, C., Yang, J., Wang, Z., Lu, X., Yang, M.H.: Diversified texture
  synthesis with feed-forward networks. IEEE Conference on Computer Vision and
  Pattern Recognition  (2017)

\bibitem{liu2017unsupervised}
Liu, M.Y., Breuel, T., Kautz, J.: Unsupervised image-to-image translation
  networks. Advances in Neural Information Processing Systems  (2017)

\bibitem{liu2015deep}
Liu, Z., Luo, P., Wang, X., Tang, X.: Deep learning face attributes in the
  wild. International Conference on Computer Vision  (2015)

\bibitem{ma2018mdcn}
Ma, W., Wu, Y., Wang, Z., Wang, G.: Mdcn: Multi-scale, deep inception
  convolutional neural networks for efficient object detection. In: 2018 24th
  International Conference on Pattern Recognition (ICPR). pp. 2510--2515. IEEE
  (2018)

\bibitem{makhzani2015adversarial}
Makhzani, A., Shlens, J., Jaitly, N., Goodfellow, I., Frey, B.: Adversarial
  autoencoders. Workshop Track of International Conference on Learning
  Representations  (2016)

\bibitem{PENG2018262}
Peng, C., Gao, X., Wang, N., Li, J.: Face recognition from multiple stylistic
  sketches: Scenarios, datasets and evaluation. Pattern Recognition
  \textbf{84},  262 -- 272 (2018)

\bibitem{radford2015DCGAN}
Radford, A., Metz, L., Chintala, S.: Unsupervised representation learning with
  deep convolutional generative adversarial networks. International Conference
  on Learning Representations  (2016)

\bibitem{ronneberger2015u}
Ronneberger, O., Fischer, P., Brox, T.: U-net: Convolutional networks for
  biomedical image segmentation. International Conference on Medical image
  computing and computer-assisted intervention pp. 234--241 (2015)

\bibitem{YU201781}
Shiqi, Y., Haifeng, C., Qing, W., Linlin, S., Yongzhen, H.: Invariant feature
  extraction for gait recognition using only one uniform model. Neurocomputing
  \textbf{239},  81 -- 93 (2017)

\bibitem{taigman2016unsupervised}
Taigman, Y., Polyak, A., Wolf, L.: Unsupervised cross-domain image generation.
  International Conference on Learning Representations  (2017)

\bibitem{tolstikhin2017wasserstein}
Tolstikhin, I., Bousquet, O., Gelly, S., Schoelkopf, B.: Wasserstein
  auto-encoders. International Conference on Learning Representations  (2018)

\bibitem{tran2017DRGAN}
Tran, L., Yin, X., Liu, X.: Disentangled representation learning gan for
  pose-invariant face recognition. IEEE Conference on Computer Vision and
  Pattern Recognition  (2017)

\bibitem{8214214}
Wang, M., Panagakis, Y., Snape, P., Zafeiriou, S.P.: Disentangling the modes of
  variation in unlabelled data. IEEE Transactions on Pattern Analysis and
  Machine Intelligence (PAMI)  \textbf{40}(11),  2682--2695 (2018)

\bibitem{wang2016generative}
Wang, X., Gupta, A.: Generative image modeling using style and structure
  adversarial networks. European Conference on Computer Vision  (2016)

\bibitem{coopnets}
Xie, J., Lu, Y., Gao, R., Zhu, S.C., Wu, Y.N.: Cooperative training of
  descriptor and generator networks. IEEE transactions on pattern analysis and
  machine intelligence (PAMI)  (2018)

\bibitem{xu2019}
Xu, W., Shawn, K., Wang, G.: Adversarially approximated autoencoder for image
  generation and manipulation. IEEE Transactions on Multimedia  (2019)

\bibitem{yi2017dualgan}
Yi, Z., Zhang, H., Gong, P.T., et~al.: Dualgan: Unsupervised dual learning for
  image-to-image translation. International Conference on Computer Vision
  (2017)

\bibitem{yu2014fine}
Yu, A., Grauman, K.: Fine-grained visual comparisons with local learning. IEEE
  Conference on Computer Vision and Pattern Recognition  (2014)

\bibitem{GaitG}
Yu, S., Liao, R., An, W., Chen, H.: Gaitganv2: Invariant gait feature
  extraction using generative adversarial networks. Pattern Recognition  (2019)

\bibitem{zhang2018unreasonable}
Zhang, R., Isola, P., Efros, A.A., Shechtman, E., Wang, O.: The unreasonable
  effectiveness of deep features as a perceptual metric. IEEE Conference on
  Computer Vision and Pattern Recognition  (2018)

\bibitem{zhang2017age}
Zhang, Z., Song, Y., Qi, H.: Age progression/regression by conditional
  adversarial autoencoder. IEEE Conference on Computer Vision and Pattern
  Recognition  (2017)

\bibitem{zhu2016generative}
Zhu, J.Y., Kr{\"a}henb{\"u}hl, P., Shechtman, E., Efros, A.A.: Generative
  visual manipulation on the natural image manifold. European Conference on
  Computer Vision  (2016)

\bibitem{CycleGAN2017}
Zhu, J.Y., Park, T., Isola, P., Efros, A.A.: Unpaired image-to-image
  translation using cycle-consistent adversarial networks. International
  Conference on Computer Vision  (2017)

\bibitem{zhu2017toward}
Zhu, J.Y., Zhang, R., Pathak, D., Darrell, T., Efros, A.A., Wang, O.,
  Shechtman, E.: Toward multimodal image-to-image translation. Advances in
  Neural Information Processing Systems pp. 465--476 (2017)

\end{thebibliography}
\end{document}